\documentclass{article}
\usepackage{arxiv}

\usepackage{graphicx}
\usepackage{amsmath}

	\usepackage[utf8]{inputenc}
	\usepackage{amsmath,amssymb,amsfonts}
	\usepackage{textcomp} 
	\usepackage[british]{babel}	
	\usepackage{csquotes}	

	\usepackage{graphicx}
	\graphicspath{{Figures/}}
	\usepackage[caption=false,font=footnotesize]{subfig}
	\usepackage{dblfloatfix} 
	
	\usepackage{array}
	\usepackage{multirow}
	\usepackage{booktabs}
	\usepackage{adjustbox}
	\usepackage{longtable}

	\usepackage{algorithm}
	\usepackage{algorithmicx}
	\usepackage[noend]{algpseudocode}
		\algrenewcommand\algorithmicindent{2.0em}%
		\algnewcommand\Let[2]{\State #1 $\gets$ #2}
		\algnewcommand\AND{\ \textbf{and}\ }
		\algnewcommand\OR{\ \textbf{or} \ }
		\algnewcommand\algorithmicinput{\textbf{Input:}}
		\algnewcommand\Input{\item[\algorithmicinput]}
		\algnewcommand\algorithmiccompute{\textbf{Compute:}}
		\algnewcommand\Compute{\item[\algorithmiccompute]}
		\algnewcommand\algorithmicoutput{\textbf{Output:}}
		\algnewcommand\Output{\item[\algorithmicoutput]}

	\usepackage{enumerate}
	\usepackage{enumitem}
	\usepackage{url}

\usepackage[backend = biber, style=apa]{biblatex}
\DeclareLanguageMapping{british}{british-apa}
\usepackage[pdftex]{hyperref}

	\newcommand{\Argmin}[1]{\ensuremath{\mathrm{Arg}\underset{#1}{\mathrm{min}\,}}}

	\DeclareMathAlphabet{\mymathbb}{U}{BOONDOX-ds}{m}{n}
	\newcommand{\unit}{\ensuremath{\mymathbb 1}}

	\def\eg{e.g.,\ }
	\def\Eg{E.g.,\ }
	\def\cf{\textit{cf.}\ }

	\newcommand{\val}[1]{\ensuremath{{#1}^{\mbox{\scriptsize Val}}}}

\begin{document}

\title{Domain Adaptation for One-Class Classification: Monitoring the Health of Critical Systems Under Limited Information}

\author{Gabriel Michau\\
ETH Zürich,\\
Swiss Federal Institute of Technology,\\
Zurich, Switzerland\\
\And 
Olga Fink,\\ 
ETH Zürich,\\
Swiss Federal Institute of Technology,\\
Zurich, Switzerland\\
}

\maketitle

\begin{abstract}
The failure of a complex and safety critical industrial asset can have extremely high consequences. Close monitoring for early detection of abnormal system conditions is therefore required.
Data-driven solutions to this problem have been limited for two reasons: First, safety critical assets are designed and maintained to be highly reliable and faults are rare. Fault detection can thus not be solved with supervised learning. Second, complex industrial systems usually have long lifetime during which they face very different operating conditions. In the early life of the system, the collected data is probably not representative of future operating conditions, making it challenging to train a robust model.

In this paper, we propose a methodology to monitor the systems in their early life. To do so, we enhance the training dataset with other units from a fleet, for which longer observations are available. Since each unit has its own specificity, we propose to extract features made independent of their origin by three unsupervised feature alignment techniques. First, using a variational encoder, we impose a shared probabilistic encoder/decoder for both units. Second, we introduce a new loss designed to conserve inter-point spacial relationships between the input and the learned features. Last, we propose to train in an adversarial manner a discriminator on the origin of the features. Once aligned, the features are fed to a one-class classifier to monitor the health of the system. By exploring the different combinations of the proposed alignment strategies, and by testing them on a real case study, a fleet composed of 112 power plants operated in different geographical locations and under very different operating regimes, we demonstrate that this alignment is necessary and beneficial.
\end{abstract}

\section{Introduction}
\label{sec:intro}

\subsection{Monitoring Complex Industrial Systems}

The industry is currently experiencing major changes in the condition monitoring of machines and production plants as data is always more easily captured and collected. This opens many opportunities to rethink the traditional problems of Prognostics and Health Management and to look for data-driven solutions, based on machine learning. Machine learning is being increasingly applied to fault detection, fault diagnostics (usually both combined in a classification task), and to the subsequent steps such as fault mitigation or decision support. One crucial component in the success of learning from the data, is the representativeness of the collected dataset used to train the models. Most modern machine learning methods, and deep learning in particular, require for their training not only a large collection of examples to learn from but also a representative sampling of these examples over the input space, or in other words, over the possible operating conditions. In fact, if the interpolation capabilities of machine learning have been acclaimed in many fields, extrapolation remains a challenge. Since most machine learning tools perform local transformation of the data to achieve better separability, it is very difficult to prove that these transformations are still relevant for data that are outside the value range used for training the models or in our case, for data stemming from different operating conditions.

This representativeness requirement on the training set is a major constraint for the health monitoring of complex or critical industrial systems, such as passenger transporting vehicles, power plants, or any systems whose failure would lead to dramatic consequences, and this for the following reasons: First, due to the fact that failures of such systems are by nature unacceptable, these systems are reliable by design and preventive maintenance is regularly performed to minimise any risk of a major fault developing. In addition, possible faults, if extremely unlikely, are plentiful and this prevents the gathering of enough data to perform data-driven fault recognition and classification. Second, the system operating conditions might evolve over very-long time scale (\eg yearly trends in a power plant operation). Collecting a representative dataset to train a reliable data-driven health monitoring model would require too much time.

These two limitations, missing faulty patterns to learn from and the need of data representative of all operating conditions, have solutions in the literature but it is seldom that both problems are considered together. First, instead of learning the faulty patterns, novelty detection methods exist, and have already been successfully applied. Yet, when data representative of all possible operating conditions are lacking, such approaches face the difficult problem of distinguishing between anomalies and a normal evolution of the system that was not observed in the training dataset. To the opposite, in the second case, many works have focused on domain adaptation, that is, either on identifying patterns significant of faults that are independent of the operating conditions or on their adaptation to new conditions. Such approaches require however examples of all possible faults in some operating conditions, to then generalise their characteristics to other operating conditions.

A intuitive approach to this domain adaptation task is to consider several similar systems each with different operating conditions and to learn fault signatures valid across all systems. This is the \textit{fleet approach} to domain transfer. A trivial context is to assume the systems identical in design and in usage, such that a single model can be trained for the whole fleet. But the task becomes more challenging when one of both constraints is relaxed. First, in a fleet from the operator perspective \parencite{Jin2015}, the units can come from different manufacturers but the units are used similarly. In this case the monitoring of the units might vary due to different sensor equipment and the challenge lies in the transformation of the data to a space independent of the sensing characteristics and technologies. Second, in a fleet from the manufacturer perspective, the units come from the same manufacturer but they are used in different operating conditions or by different operators \parencite{Leone2017}. In this second case the operating conditions will be the distinguishing elements between units and the challenge is in the alignment of the data, such that faults can be recognisable independently of the operation. Of course, the combination of both is also a possibility and would lead to an even more challenging task. In this paper, we set ourselves in the context of similarly monitored units with different operating conditions, that is, in the fleet from the manufacturer perspective. 

For the monitoring and diagnosis of such fleets, a vast literature exists proposing solutions that can be organised by increasing complexity of the task as follows:
\begin{enumerate}
    \item Identifying some relevant parameters of the units in order to adapt them to each unit or to perform clustering and use the data of each cluster to train the model. \textcite{Zio2010} compare multi-dimensional measurements, independent of time. \textcite{lapira2012fault} clusters a fleet of wind turbine based on power versus wind diagrams, pre-selecting diagrams with similar wind regimes. \textcite{Gonzalez-PriDa2016} propose an entropy inspired index based on availability and productivity to cluster the units. \textcite{Peysson2019} propose a framework for fleet-wide maintenance with knowledge base architecture, uniting semantic and systemic approach of the fleet.
    \item The entire time series are used to cluster units together. \textcite{Leone2016} compare one dimensional time series by computing the euclidean distance between a trajectory and reference trajectories. \textcite{liu2018cyber} proposes, among other, the use of time machine, clustering time series from a fleet a wind turbine with the DS3 algorithm. \textcite{al2018framework} cluster nuclear power-plant based on their shut-down transient.
    \item Model each unit functional behavior and try to identify similar ones. \textcite{Michau2018b} use the whole set of condition monitoring data to define the similarity. Such approaches do not depend on the length of the observation since the functional relationship is learnt.
    \item Align the feature space of different units, such as proposed in the present paper.
\end{enumerate}

Each level increases the complexity of the solution, but tends to mitigate some of the limitations of the previous one. The main limitations of each of the above described levels are:
\begin{enumerate}
    \item Aggregated parameters do not guarantee that all the relevant conditions have been covered. \Eg \textcite{lapira2012fault} have to first segment diagram with similar wind regimes.
    \item Comparing the distances between datasets is a problem affected by the curse of dimensionality: in high dimensions, the notion of distance loses its traditional meaning \parencite{Domingos2012}, and the temporal dimension particularly important when operating conditions evolve, make this comparison even more challenging. \Eg \textcite{al2018framework} restrict themselves to fixed-length extracted transient from the time series.
    \item Even though such approaches are more robust to variations in the behaviour of the system, sufficient similarity in the operating range is still a strong requirement, which may require large fleets for the approach to be applicable. 
    \item When the alignment is really robust to variations in the operating conditions, it can be to the point that the subsequent condition monitoring model might interpret as natural some degradation of the system and miss the alarms. 
\end{enumerate}

Aligning the feature space in the Prognostics and Health Management field is not a new idea, but it has been only applied to diagnosis or Remaining Useful Life estimation, to the best of the authors' knowledge. Such diagnostics problem have been extensively studied with traditional machine learning approaches \parencite{margolis2011literature}, but also more recently and more specifically with deep-learning \parencite{kouw2019review}. In the context of diagnostics, it is almost always assumed that the labels on the possible faults or on the degradation trajectories exist for some units, which will be used as reference for the alignment and are therefore denoted by \textit{source} units. The challenge is then to make sure that the models perform as well on the \textit{target} units for which diagnostics labels were not available in sufficient quantity or nor available at all.

Most of the alignment methods follow the same framework: First, features are extracted, engineered or learned, such as to maximise the performances of a subsequent classifier trained in the source domain where labels are available. Some works aim at adapting the target features to match the source by means of a transformation \parencite{fernando2013unsupervised,Xie2016,Zhang2017a}, others combine both alignment and feature learning in a single task. To do so, a min-max problem is solved, to minimise the classification loss in the source domain while maximising the loss of a domain discriminator. For example, \textcite{Lu2017a} train a neural network such that one intermediate layer (namely the feature or latent space) minimises the Maximum Mean Discrepancy \parencite{Borgwardt2006} between the source and the target and maximises the detection accuracy of a subsequent fault classifier. Ensuring that the origin of the features cannot be classified encourages a distribution overlap in the feature space.
\textcite{Li2018} introduced the use of generative model to transfer faults from one operating condition to another with a two level deep learning approach. Fake faulty and healthy samples in different operating conditions are generated to train in the second step, a classifier also on operating conditions where the faults were not really experienced.

As target labels are missing in the training set, such approaches are sometimes also denoted as \textit{Unsupervised Domain Adaptation}, where adaptation is performed on an unsupervised domain, that is, the target domain \parencite{fernando2013unsupervised}.
The training of the feature extractor and of the classifier is however supervised in the source domain.
Recent results demonstrate that this supervision of the feature extractor training in the source domain through the classifier is essential to the success of these approaches. By making sure that the features can be classified, the classifier constrains greatly the feature space \parencite{wang2019domain}.

These approaches cannot be directly applied in the context of unsupervised health monitoring, where anomalous labels and anomalous samples are available neither for the source nor for the target. In our context, as the training uses healthy data only, there is no classification information to back-propagate to the feature extractor. The feature extraction is in fact unsupervised with respect to the health of the system, while the alignment is still required. This setup is thus an even more challenging task never solved so far in the literature, to the best of our knowledge.

\begin{figure}
\centering
\includegraphics[width=9cm]{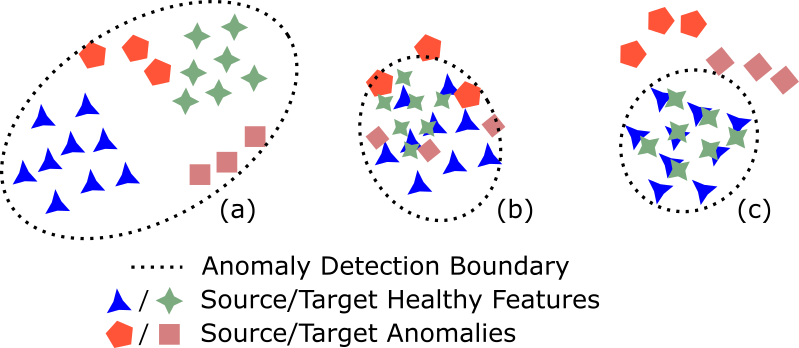}
\caption{Feature alignment: (a) Combining healthy features of source and target without alignment lead to wide anomaly detection boundaries and missed anomalies. (b) Healthy feature non-discriminability. Without constraints on the transformation, anomalies might be mixed with  healthy features. (c) By imposing non-discriminability and an homothetic transformation, inter-point relationships are kept, ensuring that the initial separability of the anomalies is conserved after the alignment.}
\label{fig:FeatAlignt}
\end{figure}

To solve this task, it seems necessary to constrain the feature space in an unsupervised manner, to convey maximal information about the health of the system within the features, as illustrated in Figure~\ref{fig:FeatAlignt}. We propose three approaches.

First, we propose an auto-encoder with shared latent space both for source and target. The features of both source and target need to be encoded in a single latent space while allowing for good input reconstruction. Using an auto-encoder is a natural extension of data-driven unsupervised anomaly detection approaches \parencite{michau_deep_2017}. We will test this approach with a variational auto-encoder, $\beta$-VAE \parencite{higgins2017beta}. VAE has shown in the past that, with its probabilistic description of the input and latent space, the obtained features are very useful for subsequent tasks \parencite{ellefsen2019remaining}.

Second, we introduce the homothety loss. The loss is designed to make sure that inter-point distance relationships are kept in the latent space for both the source and the target. If the features were obtained through an homothetic projection from the input to the feature space, they would minimise the homothety loss.

Last, we use an origin discriminator on the feature space trained in an adversarial manner. The discriminator itself is trained to best classify the origin dataset of the features while the feature extractor is trained such as the resulting features cannot be classified by the discriminator.

The remainder of the paper is organised as follows: 
Section 2 provides an overview of the tools used for the proposed approach and motivates their usage in the particular context of unsupervised feature learning and alignment. 
Section 3 presents a real application case study that faces the difficulties discussed above, including rare faults, limited observation time, limited representative condition monitoring data collected over a short period. The comparisons are performed on a fleet comprising 112 power plants monitored for one year, 12 of which experienced a fault. We limit ourselves to two months of data available for the target unit training, quite a small time interval compared to yearly fluctuations of a power plant operation.

    \subsection{Notations}
\makebox[2cm]{${\cdot}_S $} Variables for the source dataset\\
\makebox[2cm]{${\cdot}_T $} Variables for the target dataset\\
\makebox[2cm]{$X_\cdot$} Input Variable\\
\makebox[2cm]{$F_\cdot$} Feature Variable\\
\makebox[2cm]{$Y_\cdot$} One-class classifier output\\
\makebox[2cm]{$D_\cdot$} Discriminator output\\
\makebox[2cm]{$\mathcal{L}$} Loss\\
\makebox[2cm]{$\mathcal{L}_{rec}$} Reconstruction Loss\\
\makebox[2cm]{$\mathcal{L}_{KL}$} Kullback-Leibler divergence Loss\\
\makebox[2cm]{$\mathcal{L}_{FA}$} Feature Alignment Loss\\
\makebox[2cm]{$\mathcal{L}_{D}$} Discriminator Loss\\
\makebox[2cm]{$\mathcal{L}_{H}$} Homothety Loss\\
\makebox[2cm]{ELM} Extreme Learning Machine\\
\makebox[2cm]{VAE} Variational Auto-Encoder\\
\makebox[2cm]{GRL} Gradient Reversal (GR) Layer\\
\makebox[2cm]{FPR} False Positive Rate (in \%)\\

\section{Methodology for Adversarial Transfer of Unsupervised Detection}
\label{sec:ua}

    \subsection{Anomaly Detection and One-Class Classification}

The traditional framework for unsupervised fault detection usually consists in a two step approach, with first, feature extraction, and second feature monitoring. Features can stem from a physical modelling of the system, from aggregated statistics on the dataset, from varied machine learning tools or from deep learning on surrogate tasks, such as auto-encoding. The monitoring of the features can be rule based (\eg by defining a threshold on the features), or statistical (\eg $\chi^2$ or Student test) or use some machine learning tools such as clustering (K-means), nearest neighbours analysis, density based modelling, subspace analysis (\eg PCA), or one-class classification such as one-class SVM or one-class classifier neural network (see the work of \textcite{Pecht2019} for more details).
To the best of the author's knowledge, unsupervised detection has never been treated with end-to-end learning due to the lack of supervision inherent to the task.

In continuity with the traditional approaches to unsupervised fault detection, we propose here to split our architecture in a feature extractor and a feature monitoring network. To monitor the features, we propose to train a one-class classifier neural network, such as developed by \textcite{leng2015one, yan2016one, michau_deep_2017, Michau2018a}, which has proven to provide more insights on system health than the traditional state-of-the-art one-class SVM, and provides better detection performances than other indices such as the \textit{special index} \parencite{Fan2017}.

In order to handle the lack of supervision in the training of the one-class classifier, \textcite{Michau2018a} proposed to train a one-class Extreme Learning Machine (ELM). Such single layered networks have randomly drawn weights between the input layer and the hidden layer. Only the weights between the hidden layer and the output are learned. Mathematical proof has been given that they are universal approximators \parencite{huang_universal_2006}, and since the problem of finding the output weights become a single variable convex optimisation problem, there is a unique optimal solution easily found by state-of-the-art algorithms with convergence guaranties, in little time compared to the traditional training of neural networks with iterative back-propagation.

\textcite{Michau2018a} demonstrated that a good decision boundary on the output of the one-class ELM is the threshold
\begin{equation}
\label{eq:thrd}
\mbox{Thrd} = \gamma\cdot \mbox{percentile}_{p}(\vert \unit - \val{Y}\vert),
\end{equation}
where $\val{Y}$ is the ouput of the one-class classifier for a validation dataset, healthy but not used in the training. It provides an estimation of the natural fluctuation in the one-class classifier output in healthy operating conditions. $p$ represents the number of expected outliers, which is linked to the cleanness of the dataset and $\gamma$ represents the sensitivity of the detection. In this paper, we take $\gamma=1.5$ and $p=99.5$, values identified as relevant in the paper.
Outliers are discriminated from the main healthy class if they are above this threshold. 

This framework was developed and successfully applied in combination to a auto-encoder ELM (namely Hierarchical ELM or HELM) to the monitoring of a single machine for which long training was available \parencite{michau_deep_2017,Michau2018a}. In the context of short training time, the same architecture was applied to paired units (source with one year of data available and target with only two months of data available) in the work of \textcite{Michau2018b}. If this approach provided better detection rates with lower false alarm rates than using two months of data only, it faces the limitation that when the units have very different operating conditions, it is difficult to pair them such that the final model provides satisfactory performances.

In this paper, we propose to change the feature extractor from an ELM auto-encoder, that is not suitable for more advanced alignment since there is no possible back-propagation of any loss, to either a Variational Auto-Encoder (VAE) or a simple feed forward network with a new custom made loss function. We propose therefore new ways to learn features across source and target but still use the one-class classifier ELM to monitor the health of the target unit.

    \subsection{Adversarial Unsupervised Feature Alignment}
The proposed framework is composed of a feature extractor and of a one-class classifier trained with healthy data only. In order to perform feature alignment we explore three strategies in the training of the feature extractor, auto-encoding, the homothety loss and adversarial discriminator. 
These alignment strategies are not exclusive, so we also explore their different combinations.

        \subsubsection{Auto-encoding as a Feature Constraint}
An auto-encoder is a model trained to reconstruct its inputs after some transformation of the data. These transformations could be to a space of lower dimensionality (compressive auto-encoder), higher dimensionality, linear (\eg PCA) or non-linear (\eg neural networks). Auto-encoding models are a popular feature extractors as they do not require any supervision. They rely on the assumption that since the learned features can be used to reconstruct the inputs, the features should contain the most meaningful information on the data and that they will be better suitable for subsequent machine learning tasks. It is in addition quite easy to enforce some additional properties of the features that seems suitable for the task at hand such as sparsity (\eg $\ell_1$ regularisation), low coefficients (\eg $\ell_2$ regularisation), robustness to noise (denoising auto-encoder), etc... 

Among the vast family of possible auto-encoders, the variational auto-encoder, proposed by \textcite{kingma2013auto}, is learning the best probabilistic representation of the input space using a superposition of Gaussian kernels. The neurons in the latent space are interpreted as mean and variance of different Gaussian kernels, from which features are sampled and decoded. Such networks can be used as traditional auto-encoders but also as generative models: by randomly sampling the Gaussian kernels, new samples can be decoded.

The training of the variational auto-encoder consists in the minimisation of two losses, first the reconstruction loss, $\mathcal{L}_{rec}$, and second the loss on the distribution discrepancy between the input (or prior) and the Gaussian kernels, by mean of the Kullback-Leibler divergence, $\mathcal{L}_{KL}$. 
The reader interested in the implementation details of the variational auto-encoder can refer to the work of \textcite{higgins2017beta}. In this work, the concept of $\beta$-VAE is introduced and propose to apply a weight $\beta$ to the Kullback-Leibler divergence loss to either prioritise a good distribution match over a good reconstruction (particularly important when VAE is used as a generative model) or the opposite.

Variational auto-encoder have been successfully used in hierarchical architectures in many fields. In PHM, they have been used in semi-supervised learning tasks for remaining useful life prediction \parencite{yoon2017semi,ellefsen2019remaining}, and for anomaly detection \parencite{kim2018squeezed}). 

        \subsubsection{Homothety as a Feature Constraint}
An alternative to auto-encoding, is the use of more traditional feed-forward networks, on which we would impose an additional constraint on the feature space. Instead of the combined loss on both the reconstruction from the feature and on the feature distribution, we propose to introduce here a loss that encourages inter-point relationship conservation in the feature space. To do so, we define the homothety loss to keep constant the inter-point distance ratios between the input $X$ and the feature space $F$. The intuition lies on the idea that a good alignment of the two dataset should correspond to both source and target sharing the same lower dimensionality feature space while being scaled in similar ways.

The proposed homothety loss is defined as:
\begin{equation}
\mathcal{L}_H = \sum_{S\in \left\lbrace \substack{\mbox{Source}\\\mbox{Target}} \right\rbrace} \frac{1}{\vert S \vert}  \sum_{(i,j)\in S}\left\Vert \left\Vert X_i - X_j \right\Vert_2 - \eta \left\Vert F_i - F_j \right\Vert_2 \right\Vert_2
\end{equation}
where
\begin{equation}
\eta = \Argmin{\tilde{\eta}} \mathcal{L}_H (\tilde{\eta})
\end{equation}

        \subsubsection{Domain Discriminator}
For both the VAE and the Homothetic Feature Alignment, the alignment can be helped further with the help of an origin discriminator trained in an adversarial manner. This training is done by solving a min-max problem where the discriminator is trained at minimising the discrimination loss on sample origins, while the feature extractor is trained to maximise this loss, that is, to make the feature indistinguishable from the discriminator perspective.

Such adversarial training has been greatly simplified since the introduction of the Gradient Reversal Layer trick, proposed by \textcite{Ganin2016}. This simple yet efficient idea consists in connecting the feature extractor to the discriminator through an additional layer which performs the identity operation in the forward pass but negates the gradient in the backward pass. The gradient passed to the feature extractor during the backward pass goes therefore in the opposite direction as the minimisation problem would require.

For the discriminator, we propose to experiment with a classic densely connected softmax classifier, and with a Wasserstein discriminator. This setup is inspired from the Wasserstein GAN \parencite{arjovsky2017wasserstein}, where a generative model is trained to minimise the Wasserstein distance between generated samples and true samples, as to make their distribution indistinguishable. The authors demonstrate that, using the Kantorovich-Rubinstein duality, this problem can also be solved by adversarial training with a neural network playing the role of the discriminator aiming at maximising the following loss:

\begin{equation}
\mathcal{L}_{D} = \mathbb{E} \left( \mbox{disc}\left( F_{\mbox{Source}}\right)\right) - \delta_w \mathbb{E} \left(\mbox{disc}\left( F_{\mbox{Target}}\right) \right)
\end{equation}

Our implementation of the Wasserstein adversarial training takes into account the latest improvements, including the gradient penalty as proposed in \textcite{gulrajani2017improved} to ensure the requirement of the Kantorovich-Rubinstein duality that the function disc() is 1-Lipschitz, and the asymmetrically relaxed Wasserstein distance as proposed by \textcite{wu2019domain}. This relaxation, here when $\delta_w >1.0$, encourages the target feature distribution to be fully contained in the source feature distribution, but does not require full reciprocal overlap. This is important here since we assume a small non-representative dataset for the target training. It is in the nature of this task to assume that the source feature distribution will have a larger support, from which the health monitoring model can learn from.

    \subsection{Summary and overlook of final architectures}
         \subsubsection{Tested Architectures}

In summary, we compare the various propositions alone and combined together as follows:
\begin{itemize}
\item \textbf{$\beta$-VAE}: the traditional $\beta$-VAE whose features are used as input to a one-class classifier. 
\item \textbf{$\beta$-VAEDs}: The same $\beta$-VAE and one-class classifier with the addition of a softmax discriminator connected to the feature space with a GR-layer
\item \textbf{$\beta$-VAEDw}: Similar as before but with a Wasserstein discriminator. In this version the adversarial training aims at minimising the Wasserstein distance between source and target.
\end{itemize}
Similarly, for the homothetic alignment, we explore the following combinations:
\begin{itemize}
\item \textbf{HFA}: Homothetic Feature Alignment, made out of a feature extractor trained with the homothetic loss and a one-class classifier.
\item \textbf{HAFAs}: Homothetic and Adversarial Feature Alignment with softmax discriminator, similar as before with in addition a softmax discriminator connected through a GR-layer.
\item \textbf{HAFAw}: Homothetic and Adversarial Feature Alignment with a Wasserstein discriminator.
\item \textbf{AFAs}: Similar as the HAFAs architecture but without the homothetic loss.
\item \textbf{AFAw}: Similar to the HAFAw architecture but without the homothetic loss.
\end{itemize}

\begin{figure*}
\hfil
\subfloat[]{\includegraphics[width=7.5cm]{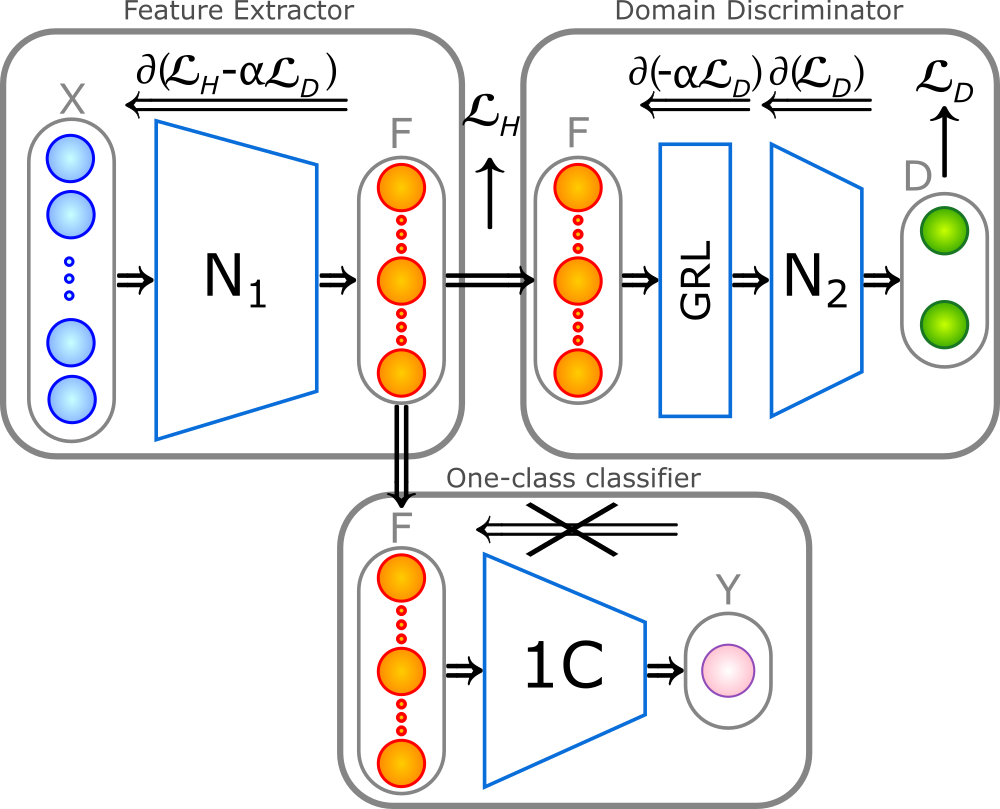}\label{sfig:HAFA}}
\hfil
\subfloat[]{\includegraphics[width=7.5cm]{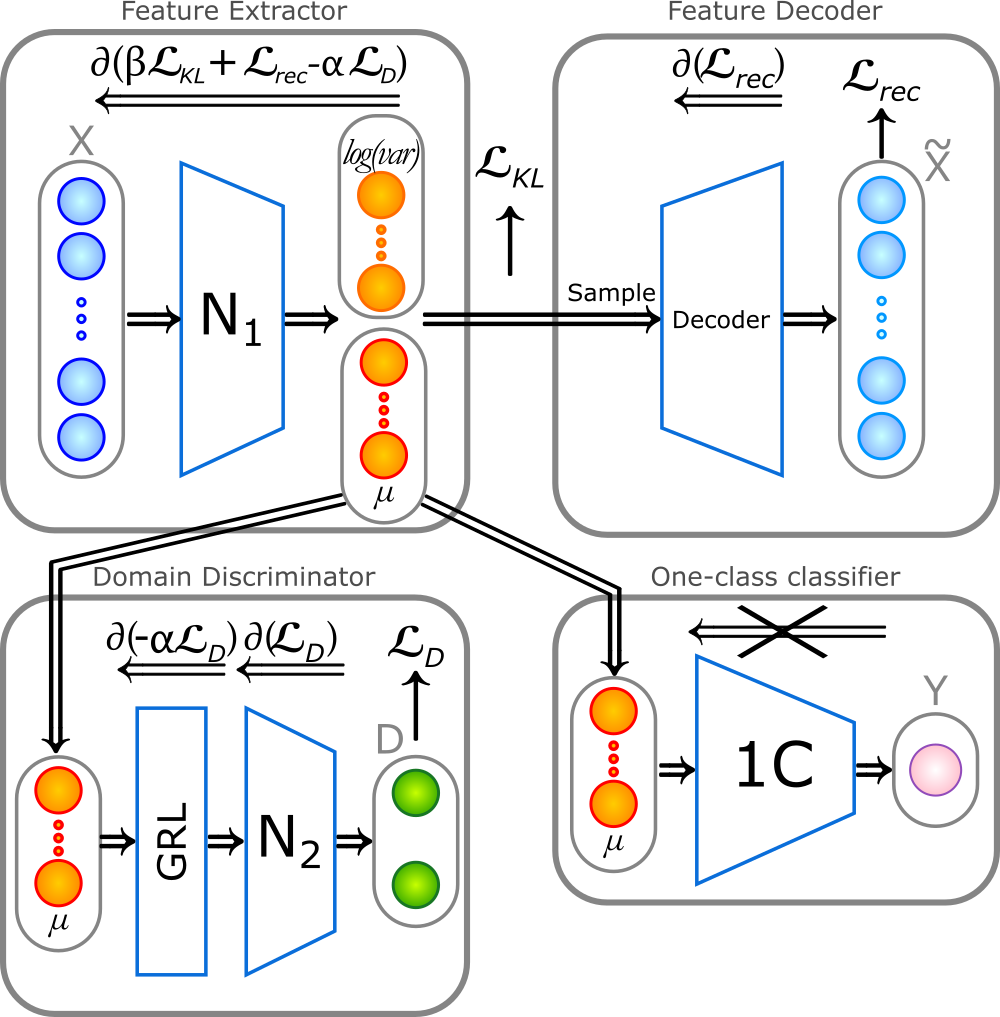}\label{sfig:VAED}}
\hfil
\caption{\textbf{Adversarial \& Unsupervised Feature Alignment Architectures}. (a) HAFA's architecture. (b) $\beta$-VAE based architectures.
The feature encoder $N_1$ is trained to minimise a feature alignment loss $\mathcal{L}_{FA}$ composed of the reversed discriminator loss $-\alpha\mathcal{L}_D$ and either (a) the homothety loss $L_H$ or (b) the variational autoencoder loss ($\mathcal{L}_{rec}+\beta\mathcal{L}_{KL}$). The discriminator $N_2$ is trained to minimise the classification loss $\mathcal{L}_D$ on the origin of the data (source vs. target). For HFA and $\beta$-VAE, the discriminator is removed. Alternatively, this corresponds to setting arbitrarily $\mathcal{L}_D=0$.}
\label{fig:UAFA}
\end{figure*}

\begin{figure}
\centering
\includegraphics[width=0.7\columnwidth]{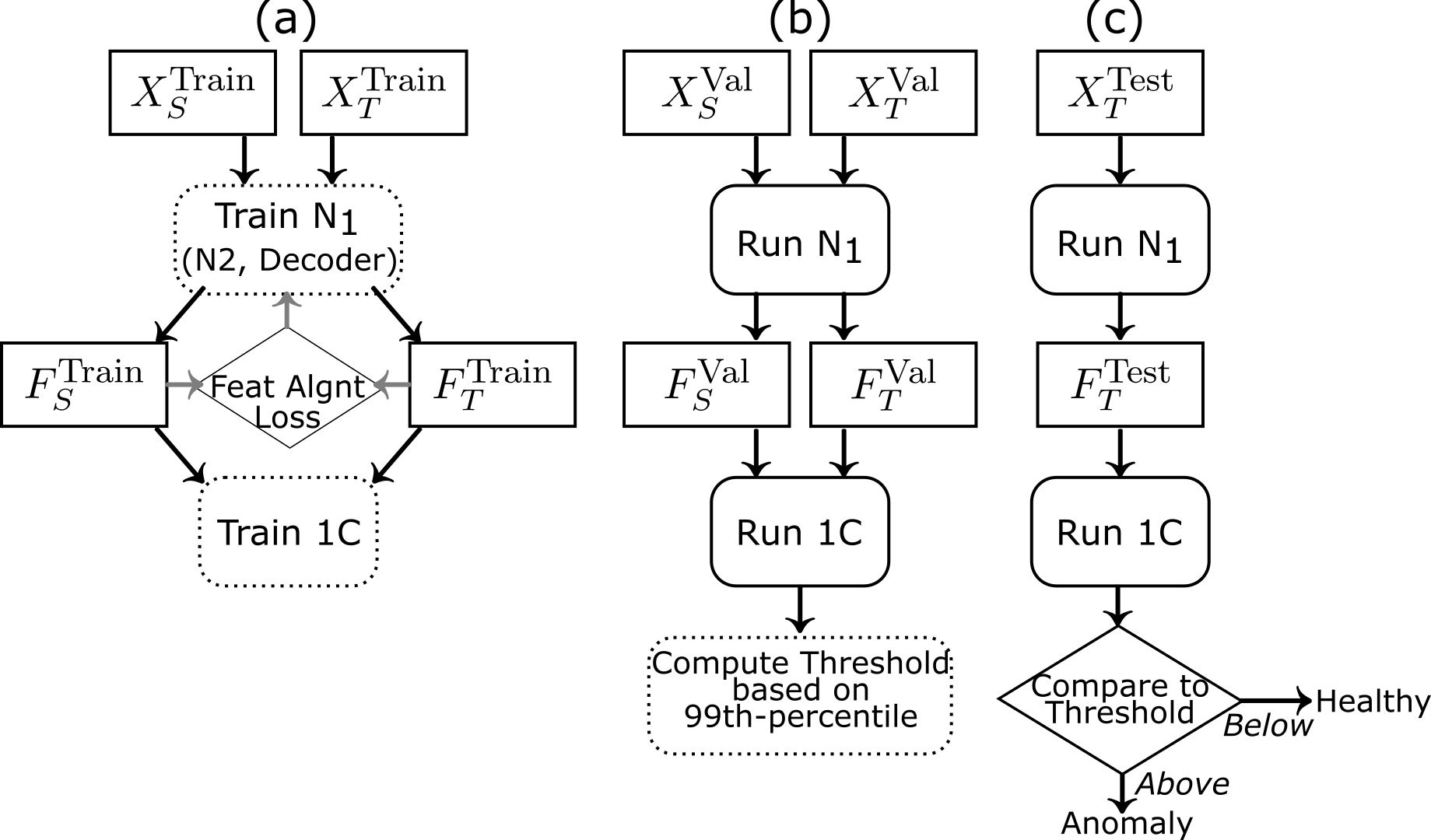}
\caption{\textbf{Flow-Chart.} (a) Using Source and Target training data, both $N_1$ and $1C$ are trained. The training of $N_1$ requires incidentally the training of the discriminator $N_2$ and of the decoder when used. (b) The output of $1C$ is analysed with a validation set from both source and target and the threshold~\eqref{eq:thrd} is computed. (c) This threshold is used as decision rule for test target data.}
\label{fig:framework}
\end{figure}

Figure~\ref{fig:UAFA} summarises the architectures explored here and Figure~\ref{fig:framework} how the whole framework is organised.

We compare in addition the results to our previous results in which units were paired together to train an HELM without alignment of any kind \parencite{Michau2018b}.

		\subsubsection{Hyper-parameters}
We tested the architecture in similar context: A two layer feature extractor with 10 neurons, a two layer discriminator with 10 and 5 neurons, \textit{relu} as activation function (but for the softmax output layer of the discriminator), a learning rate of $10^{-4}$, the ADAM optimiser \parencite{kingma2014adam}, over 200 epoch with batch size of 1000. The Wasserstein discriminator has a gradient penalty weight of 10 as proposed in the seminal paper. The VAE decoder has a symmetric architecture to that of the feature extractor.

The asymmetric coefficient of the Wasserstein discriminator $\delta_w$ is tested with $1.0$ (no asymmetric relaxation) and $4.0$ (as proposed in the seminal paper). The gradient reversal layer is tested with weights $\alpha$ set to $1.0$ and $0.2$. With $0.2$, the discriminator would be trained with a gradient 5 times higher than the feature extractor, increasing the relative training of the discriminator. The $\beta$ of the $\beta$-VAE is tested with $10.0$ (more weight to the Kullback-Leibler loss), $1.0$ and $0.1$ (more weight to the reconstruction loss). Results are only reported for $\alpha=\beta=1.0$ as all other combinations, in our setting, were providing worse performance.

\section{Case Study}
\label{sec:cs}
	\subsection{Introduction to the Case Study}

To demonstrate the suitability and effectiveness of the proposed approaches and compare between the different strategies, a comparison is performed on a fleet comprising 112 power plants, similarly to that presented in \textcite{Michau2018b, Michau2019}. 
In the available fleet, 100 gas turbines have not experienced identifiable faults during the observation period (approximately one year), they are therefore considered here as healthy and 12 units have experienced a failure of the stator vane. 

A vane in a compressor redirects the gas between the blade rows, leading to an increase in pressure and temperature.
The failure of a compressor vane in a gas turbine is usually due to a Foreign Object Damage (FOD) caused by a part loosening and travelling downstream, affecting subsequent compressor parts, the combustor or the turbine itself.
Fatigue and impact from surge can also affect the vane geometry and shape and lead to this failure mode. Parts are stressed to their limits to achieve high operational efficiency with complex cooling schemes to avoid their melting, especially during high load.
Such failures are undesirable due to the associated costs, including repair costs and operational costs of the unexpected power plant shutdown. 

Because of the various different factors that can contribute to the source of the failure mode, including assembly, material errors, or the result of specific operation profiles, the occurrence of a specific failure mode is considered as being random. Therefore, the focus is nowadays on early detection and fault isolation and not on prediction.

So far, the detection of compressor vane failures mainly relied on analytic stemming from domain expertise. 
Yet, if the algorithms are particularly tuned for high detection rates, they often generate too many false alarms. 
False alarms are very costly, each raised alarm is manually verified by an expert which makes it a time- and resource-consuming task.

	\subsection{The dataset}

The turbines are monitored with 24 parameters, sampled every 5 minutes over 1 year. They stem from 15 real sensors and 9 ISO variables (measurements modified by a physical model to represent some hand-crafted features in standard operating conditions 15$^o$ C, 1 atmosphere).
Available ISO measurements are, the power, the heat rate, the efficiency and indicators on the compressor (efficiency, pressure ratio, discharge pressure, discharge temperature, flow). Other measurements are pressures and temperatures from the different parts of the turbine and of the compressor (inlet and bell mouth), ambient condition measurements and operating state measurements such as the rotor speed, the turbine output, and fuel stroke. The data available in this case study is limited to one year, over which the gas turbines have not experienced all relevant operating conditions. We aim at being able to propose condition monitoring methods that rely on two months of data only from the turbine of interest.

To test the model and report the results, we apply the proposed methodology to the 12 gas turbines with faults as target, from which we extract the first two months of data as training set (around 17\,000 measurements). The 100 remaining gas turbines are candidate source datasets, considered as healthy. 
For the 12 target gas turbines, all data points after the first two months and until a month before the expert detected the fault (around 39\,000 measurements), are considered as healthy and are used to quantify the percentage of false positives (FPR). The last month before the detection is ignored as faults could be the consequence of prior deterioration and a detection could not be reliably compared to any ground truth. Last, the data points after the time of detection by the expert are considered as unhealthy. As many of the 12 datasets have very few points available after that detection time (from 8 to 1000 points), we will consider the fault as detected if the threshold is exceeded at least twice successively.

The validation dataset is made by extracting 6\% of the training dataset. The data has been normalised such that all variables 1st and 99th-percentiles are respectively $-1$ and $1$, such that the resulting normalisation is independent to the presence of outliers. Rows with any missing values or 0 (which is not a possible value for any of the measurement) have been removed. 

	\subsection{Alignment abilities}
\begin{table}
\centering
\setlength\tabcolsep{3pt}
\begin{adjustbox}{max width=\columnwidth}

\begin{tabular}{l|rrrrrrrrr}
\toprule
\small
Unit  &\small HELM &\small $\beta$-VAE &\small $\beta$-VAEs &\small $\beta$-VAEw &\small HFA &\small AFAs &\small AFAw &\small HAFAs &\small HAFAw \\
\midrule
1     & 11   & 74          & 65           & 74           & 83  & 83   & 85   & 86    & 79    \\
2     & 0    & 5           & 20           & 13           & 10  & 13   & 24   & 5     & 12    \\
3     & 10   & 28          & 22           & 22           & 21  & 23   & 30   & 32    & 34    \\
4     & 17   & 30          & 21           & 32           & 54  & 55   & 54   & 52    & 49    \\
5     & 94   & 68          & 47           & 67           & 90  & 59   & 63   & 80    & 85    \\
6     & 92   & 51          & 68           & 63           & 85  & 77   & 79   & 92    & 93    \\
7     & 0    & 13          & 29           & 24           & 29  & 45   & 31   & 34    & 26    \\
8     & 95   & 40          & 42           & 43           & 67  & 61   & 63   & 65    & 58    \\
9     & 2    & 19          & 19           & 18           & 26  & 28   & 32   & 22    & 39    \\
10    & 1    & 18          & 15           & 8            & 21  & 28   & 24   & 34    & 29    \\
11    & 2    & 20          & 35           & 47           & 59  & 63   & 51   & 60    & 51    \\
12    & 0    & 3           & 3            & 4            & 2   & 2    & 1    & 1     & 3     \\
\midrule
\small R\% (5\%) & 27.3  & 31.1     & 32.5          & 34.9         & 46.0 & 45.2  &45.2  & 47.4   & 47.0  \\
\small R\% (1\%) & 13.5 & 20.6 & 22.0 & 25.8 & 30.5 & 27.0 & 25.5 & 32.8 & 30.1\\
\bottomrule
\end{tabular}
\end{adjustbox}\caption{Number of successfully aligned pairs (detected fault and less than 5\% FPR). Last rows R are the mean ratio of aligned pairs (in \%), with selection threshold on FPR at 5\% and 1\%.}
\label{tbl:alipair}
\end{table}

\begin{table}
\centering
\setlength\tabcolsep{3pt}
\begin{adjustbox}{max width=\columnwidth}
\begin{tabular}{l|rrrcrrrrr}
\toprule
\small
Unit &\small HELM  &\small $\beta$-VAE &\small $\beta$-VAEs &\small $\beta$-VAEw &\small HFA  &\small AFAs &\small AFAw &\small HAFAs &\small HAFAw \\
\midrule
1    & 1.46  & 0.00        & 0.01         & 0.04         & 0.00 & 0.00 & 0.04 & 0.01  & 0.00  \\
2    & 10.00 & 0.12        & 0.13         & 0.00         & 0.01 & 0.05 & 0.03 & 0.52  & 0.02  \\
3    & 0.35  & 0.01        & 0.18         & 0.31         & 0.10 & 0.01 & 0.00 & 0.04  & 0.49  \\
4    & 0.81  & 0.08        & 0.00         & 0.00         & 0.01 & 0.02 & 0.01 & 0.06  & 0.01  \\
5    & 0.00  & 0.00        & 0.00         & 0.00         & 0.00 & 0.00 & 0.00 & 0.00  & 0.00  \\
6    & 0.35  & 0.01        & 0.03         & 0.00         & 0.02 & 0.02 & 0.01 & 0.02  & 0.01  \\
7    & 6.45  & 1.85        & 0.09         & 0.27         & 0.15 & 0.02 & 0.17 & 0.62  & 0.23  \\
8    & 0.00  & 0.00        & 0.00         & 0.00         & 0.00 & 0.00 & 0.00 & 0.00  & 0.00  \\
9    & 3.20  & 0.14        & 0.09         & 0.06         & 0.05 & 0.01 & 0.03 & 0.09  & 0.00  \\
10   & 4.91  & 0.20        & 0.07         & 0.53         & 0.25 & 0.07 & 0.29 & 0.21  & 0.72  \\
11   & 4.71  & 0.15        & 0.01         & 0.10         & 0.09 & 0.08 & 0.00 & 0.01  & 0.15  \\
12   & 7.27  & 1.61        & 1.41         & 1.76         & 2.94 & 1.01 & 3.37 & 2.64  & 2.65  \\
\midrule
Mean & 3.29  & 0.35        & 0.17         & 0.26         & 0.30 & 0.11 & 0.33 & 0.35  & 0.36 \\
\bottomrule
\end{tabular}
\end{adjustbox}
\caption{FPR for the best models.}
\label{tbl:bestmodel}
\end{table}

The results presented in this section aim at comparing the different combinations, and highlight the benefits of each components, in combination of others or alone. To provide insights on how well each model is able to align features from the different units of the fleet, we paired each of the 12 units, considered as target, with every one of the 100 healthy units, considered as source. 

The fleet of tested unit has a strong variability and most pairs will not provide an interesting model. Therefore, a good indicator to compare the methodologies in their capability to take benefit of a pair of datasets is to tally the number of pairs leading to a relatively successful model, that is, a model that detect the fault in the target, and with a low FPR. Thus we report the results by setting an arbitrary threshold at 5\% FPR. We also report the aggregated results with a threshold at 1\%, to demonstrate that the comparative study conclusions remain valid independently of this selection process.
Out of the resulting 34 combinations (8 architectures and different hyper-parameter settings) trained and tested on the 1200 pairs, we report the results for each architecture for the hyper-parameters maximising the overall number of successfully models.

We report in Table~\ref{tbl:alipair} for each unit, how many aligned pairs were achieved with each combination. 
We also report in Table~\ref{tbl:bestmodel} the lowest FPR achieved for each of the 12 units, for models which could detect the fault.
We ran the experiments with Euler V\footnote{\url{https://scicomp.ethz.ch/wiki/Euler}}, with two cores from a processor Intel Xeon E5-2697v2. The heaviest models (HAFA's family) took around two minutes to be trained and tested. 40800 models were trained totalling more than 1.5 months of computation time.

\section{Discussion}
\label{sec:disc}

\begin{table}
\centering
\setlength\tabcolsep{3pt}
\begin{adjustbox}{max width=\columnwidth}
\begin{tabular}{l|rrrrrrrrrr}
\toprule
\small Unit & \tiny{2mHELM} &\small HELM  &\small $\beta$-VAE &\small $\beta$-VAEs &\small $\beta$-VAEw &\small HFA  &\small AFAs  &\small AFAw  &\small HAFAs      &\small HAFAw      \\
\midrule
1            & 9.05    & 1.46  & 0.98        & 0.55         & 0.74         & 3.61 & 0.32  & 4.77  & 0.52       & 0.82       \\
2            &         & 15.12 & 8.55        & 7.88         & 0.34         &      &       &       &            & 5.48       \\
3            & 18.90   & 13.22 & 4.89        & 0.70         & 4.63         &      &       &       & 4.54       & 3.05       \\
4            & 42.61   & 26.34 &             &              &              & 1.51 & 2.18  & 0.70  & 3.81       &            \\
5            & 4.20    & 1.50  & 0.08        & 3.48         & 0.08         &      & 0.01  & 0.03  & 1.02       & 0.26       \\
6            & 3.14    & 1.16  & 1.16        & 1.25         & 0.66         & 2.13 & 1.14  & 0.27  & 1.71       & 0.99       \\
7            & 56.55   & 20.33 & 13.30       & 1.26         & 14.17        & 5.43 &       & 6.22  & 0.62       & 4.77       \\
8            & 0.19    & 0.08  &             & 1.28         &              & 0.03 &       & 0.19  & 0.05       & 0.16       \\
9            & 20.13   &       & 13.11       & 6.65         & 0.06         & 4.56 &       &       & 0.63       & 0.20       \\
10           & 81.76   & 39.79 & 2.26        &              & 8.06         &      & 0.07  & 30.62 &            & 2.71       \\
11           & 36.18   & 25.53 & 14.72       & 0.15         & 3.09         & 2.97 & 0.26  & 0.90  & 4.37       & 0.17       \\
12           & 68.60   & 37.05 &             &              &              &      & 19.60 & 8.10  &            &            \\
\midrule
\#AP & 3       & 4     & 5           & 7            & 7            & 6    & 6     & 6     & \textbf{9} & \textbf{9} \\
\bottomrule
\end{tabular}
\end{adjustbox}
\caption{FPR for the pairs minimising the Maximum Mean Discrepancy. Empty cells correspond to models with missed fault. The last row (\#AP) contains the number of models with less than 5\% FPR.}
\label{tbl:mmdsel}
\end{table}

The results presented in Table~\ref{tbl:alipair} demonstrate that the proposed alignment methodology all bring positive impact on the problem of transferring operating conditions between units for unsupervised health monitoring of the gas turbines. Compared to the naive combination of the source and target dataset in a single training set with HELM, all proposed alignment methods improve the number of successfully aligned pairs (target fault detected and less than 5\% FPR). Each alignment strategy improves the results and leads to very efficient models as illustrated in Table~\ref{tbl:bestmodel} (most aligned models have far less than 1\% FPR). The homothetic loss is clearly the most contributing factor to the alignment as all architectures making use of it align successfully more pairs. The use of adversarial discriminator is also contributing quite significantly to the alignment process. It improves the results from the $\beta$-VAE significantly and provides good alignment even when used alone (\cf AFAs and AFAw). When used with the $\beta$-VAE, the Wasserstein discriminator provides the best results with an asymmetric coefficient $\delta_w=4.0$. This demonstrates the importance to relax the distribution alignment. In the homothetic alignment conditions, the selected asymmetric coefficient is $\delta_w=1.0$, showing that the homothetic loss encourages already a distribution overlap and reduces the need of asymmetric alignment. In that case, the classic softmax discriminator is actually providing better results.

The results presented above demonstrate that, first the alignment procedures lead to an increased probability of training a useful model given a pair of units and that second, the models are performing better with alignment (lower FPR and higher detection rate). An interesting question is the \textit{a priori} selection of the pair of units which, once aligned, can be used to train a successful health monitoring model. While this question is left for future research, a possible solution is to compare units based on few relevant characteristics. Here we propose a possible solution consisting in the selection of the pair for which the Maximum Mean Discrepancy (MMD) on two representative variables is minimal (Output Power and Inlet Guide Vanes Angle). The results are reported in Table~\ref{tbl:mmdsel}. Empty cells corresponds to cases where the fault was not detected by the model. For comparison purposes, these models are compared to a simple HELM only trained with the two months of data from the target unit (2mHELM). Based on this imperfect selection process, the proposed alignment procedures all improves the results, both in number of useful models but also in reducing the FPR. Previous results demonstrated that the units are very different from each others, this is highlighted here with the relatively low number of successfully aligned pairs for units 2, 3, 7, 9, 10, 11 and 12. These units have also very high variability in their operating conditions as shown by the very high FPR of the model trained only on their first two months of data. This high variability makes it extremely challenging to identify the other units likely to bring beneficial information to the model. 

To the opposite, units 5, 6 and 8 looks very stable in their operation as the models trained only on the first two months already provide satisfactory results (less than 5\% FPR). Already with HELM they could be matched with almost all other units. This can explain that aligning those unit with other sources is not necessary and might actually confuse the subsequent one-class classifier (for these units, the number of successfully aligned pairs decreases with some of the alignment procedures). Yet, for the best performing approaches (HAFAs and HAFAw), the results are improved, even on these units.

Also, the results presented in this paper focused on providing a fair comparison of different alignment strategies. The different combinations trained on the 1200 pairs led to the training of over 40\,000 models. Once a strategy is chosen, plenty of room is left for hyper-parameter tuning, which can only improve the results presented here.

\section{Conclusions}

In this paper, we tackled the problem of feature alignment for unsupervised anomaly detection. In the context where a fleet of units is available, we proposed to align a target unit, for which little condition monitoring data are available, with a source unit, for which longer observations period had been recorded, both in healthy conditions. Contrarily to the traditional case in domain alignment, the feature learning cannot rely on the back-propagation of the loss of a subsequent classifier. Instead, we presented three alignment strategies, auto-enconding in shared latent space, the newly proposed homothety loss and the adversarial training of a discriminator (with a traditional softmax classifier and a Wasserstein discriminator). All strategies improve the results for the subsequent unsupervised anomaly detection model. Among these strategies, we demonstrated that the newly proposed homothety loss has the strongest impact on the results and can be further improved by the use of a discriminator.

In the future, deeper analysis of the unit characteristics might help to identify the units for which additional data is required and to also identify which other units to select among the fleet. Such analysis could rely on expert knowledge or on an online adaptation of the sourcing data depending on the current operation. Such approach would face the challenge of distinguishing new from degraded operating conditions. Another interesting trail is the selection of the sourcing data among the whole fleet rather than attempting to pair specific units. In such case, the right data selection remains an open research question.

\section*{Acknowledgement}
This research was funded by the Swiss National Science Foundation (SNSF)
Grant no. PP00P2 176878.
\printbibliography
 
\end{document}